\documentclass{article}

\usepackage[preprint]{neurips_2026}
\makeatletter\renewcommand{\@notice}{}\makeatother

\usepackage[utf8]{inputenc}
\usepackage[T1]{fontenc}
\usepackage{hyperref}
\usepackage{url}
\usepackage{booktabs}
\usepackage{amsmath}
\usepackage{amsfonts}
\usepackage{amssymb}
\usepackage{nicefrac}
\usepackage{microtype}
\usepackage{xcolor}
\usepackage{graphicx}
\usepackage{tikz}
\usetikzlibrary{positioning, arrows.meta, fit, calc, backgrounds, shapes.geometric}
\usepackage{eso-pic}

\title{MOSS: Self-Evolution through Source-Level Rewriting in Autonomous Agent Systems}

\author{%
  Qianshu Cai\textsuperscript{1,\,$\ast$,\,$\triangle$}\enspace
  Yonggang Zhang\textsuperscript{2,\,$\ast$}\enspace
  Xianzhang Jia\textsuperscript{2}\enspace
  Huajiang Zheng\textsuperscript{2} \\
  \bfseries\rule{0pt}{24pt}%
  Wei Xue\textsuperscript{2}\enspace
  Jun Song\textsuperscript{3}\enspace
  Xinmei Tian\textsuperscript{1,\,$\dagger$}\enspace
  Yike Guo\textsuperscript{2,\,$\dagger$} \\[6pt]
  \textsuperscript{1}University of Science and Technology of China \\
  \textsuperscript{2}Hong Kong Generative AI Research \& Development Center \\
  \textsuperscript{2}The Hong Kong University of Science and Technology \\
  \textsuperscript{3}Hong Kong Baptist University
}
\begin{document}

\AddToShipoutPictureFG*{%
  \AtPageUpperLeft{%
    \put(\LenToUnit{1.5in},-\LenToUnit{0.85in}){%
      \includegraphics[height=1.2cm]{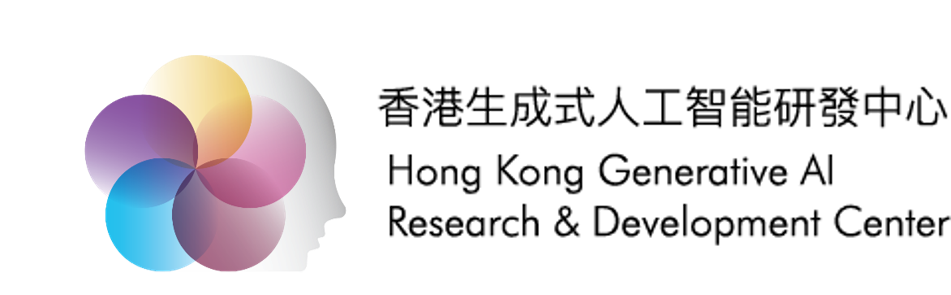}%
    }%
  }%
}

\maketitle
 
\renewcommand{\thefootnote}{$\ast$}\footnotetext{Equal contribution.}
\renewcommand{\thefootnote}{$\dagger$}\footnotetext{Corresponding authors.}
\renewcommand{\thefootnote}{$\triangle$}\footnotetext{Remote RA at HKGAI and HKUST.}
\renewcommand{\thefootnote}{\arabic{footnote}}

\begin{abstract}
Autonomous agentic systems are largely static after deployment: they do not learn from user interactions, and recurring failures persist until the next human-driven update ships a fix. Self-evolving agents have emerged in response, but all confine evolution to text-mutable artifacts---skill files, prompt configurations, memory schemas, workflow graphs---and leave the agent harness untouched. Since routing, hook ordering, state invariants, and dispatch live in code rather than in any text artifact, an entire class of structural failure is physically unreachable from the text layer. We argue that source-level adaptation is a fundamentally more general medium: it is Turing-complete, a strict superset of every text-mutable scope, takes effect deterministically rather than through base-model compliance, and does not erode under long-context drift. We present MOSS, a system that performs self-rewriting at the source level on production agentic substrates. Each evolution is anchored to an automatically curated batch of production-failure evidence and proceeds through a deterministic multi-stage pipeline; code modification is delegated to a pluggable external coding-agent CLI while MOSS retains stage ordering and verdicts. Candidates are verified by replaying the batch against the candidate image in ephemeral trial workers, then promoted via user-consent-gated, in-place container swap with health-probe-gated rollback. On OpenClaw, MOSS lifts a four-task mean grader score from 0.25 to 0.61 in a single cycle without human intervention. GitHub: \url{https://github.com/hkgai-official/Moss}.
\end{abstract}

\section{Introduction}
\label{sec:intro}

Application-level autonomous agent systems such as OpenClaw~\citep{openclaw-github} have matured from research demonstrations into real-world workers, deployed across Slack, Discord, web, and other channels to complete complex multi-step tasks for users. Yet once deployed, they remain largely static: they do not learn from how they are actually used, and the same failure modes recur across users until the next human-driven iteration ships a fix. A wave of self-evolving agents has emerged in response, letting agents evolve themselves after deployment~\citep{hermes-github, hermes-self-evo, wang2026procedural, ma2026skillclaw, liang2026genericagent, wang2025evoagentx}.

These systems leave one layer untouched. As Table~\ref{tab:scope-comparison} makes visible, their evolution scope is bounded to \textbf{text-mutable artifacts}---skill files, prompt configurations, memory schemas, and at most workflow graphs; the \textbf{agent harness}---routing, state management, dispatch, hooks, mediator, session lifecycle---is never modified by the agent itself. The limit this boundary imposes is physical: edits to text-mutable artifacts can only change \emph{what the agent itself thinks} (what to say, which skill to invoke, how to break a task into sub-goals); they cannot change \emph{what the harness decides on its behalf}. Once a failure originates in this layer---mis-routed messages, hooks firing out of order, corrupted session state, atomicity bugs across concurrent skills---no update to skills, prompts, or memory can reach it: the bug is not in the prompt text, and a prompt rewrite cannot paper over it. The proportion of failures of this kind scales with harness complexity, so this gap widens as agentic systems mature.

\begin{table}[ht]
\caption{Scope of evolution across application-level self-evolving agentic systems.}
\label{tab:scope-comparison}
\centering
\small
\begin{tabular}{lcccc}
\toprule
Project & Skill & Prompt & Memory & Harness \\
\midrule
Hermes Agent~\citep{hermes-github}     & $\checkmark$ & $\times$              & $\checkmark$ & $\times$ \\
SkillClaw~\citep{ma2026skillclaw}            & $\checkmark$ & $\times$              & $\times$     & $\times$ \\
GenericAgent~\citep{liang2026genericagent}             & $\checkmark$ & $\times$              & $\checkmark$ & $\times$ \\
EvoAgentX~\citep{wang2025evoagentx}                 & $\checkmark$ & $\checkmark$          & $\times$     & $\times$ \\
\textbf{MOSS (Ours)}                             & \boldmath$\checkmark$ & \boldmath$\checkmark$ & \boldmath$\checkmark$ & \boldmath$\checkmark$ \\
\bottomrule
\end{tabular}
\end{table}

We argue that source-level adaptation---where an agent performs self-rewriting on its own source---is a fundamentally more general medium, simultaneously superior to text-mutable evolution along four axes. It is Turing-complete: because programming languages are Turing-complete, the source-code design space constitutes a universal search space within which every text-mutable agent-design space---prompts, skills, memory schemas, workflow graphs---sits as a strict subset~\citep{hu2025automated}; editing code can reach not only every configuration these subsets can represent, but arbitrary agentic structures they cannot. It is a strict superset of every text-mutable scope: whatever a prompt edit can achieve, an equivalent code edit can also achieve, and not the other way around. It takes effect deterministically: routing logic, hook ordering, and state-machine invariants run as code, and their behavior does not depend on whether the base model correctly reads new text and complies with it---text-mutable fixes are the opposite, with effectiveness rising and falling with current base-model capability. And it does not erode under long-context drift: text-mutable fixes are prompts, skills, and memory entries that the agent must re-read on every turn, and as these accumulate over weeks of production use the model's adherence to any single piece of guidance dilutes; edits at the source layer are encoded as behavior, not text to be re-read, and so do not degrade as the system ages.

Academic work has shown source-level self-rewriting to be feasible on minimal scaffolds: SICA~\citep{robeyns2025self}, the Darwin G\"{o}del Machine~\citep{zhang2025darwin}, and HyperAgents~\citep{zhang2026hyperagents} all demonstrate an agent modifying its own implementation to lift its own benchmark scores; Meta-Harness~\citep{lee2026meta} further shows that exposing past execution traces to the coding-agent proposer drives iterative improvement more strongly than benchmark scores alone do. But these systems all run on minimal scaffolds with feedback gated by benchmark scores---fundamentally an exploratory paradigm. Carrying the same source-level capability over to an application-level, production-grade substrate is a different engineering setting altogether: the codebase is large, there is no clean benchmark score to anchor evolution against, and live user traffic together with persistent user state must survive. This pushes a series of concrete questions up to the system-design layer: when to trigger an evolution attempt, where in the codebase the fix should land, whether the resulting candidate is actually better than the deployed version, and how to put that candidate in front of real users without disrupting the live deployment.

We present \textbf{MOSS}, a system that performs comprehensive source-level self-rewriting on production agentic substrates---an instantiation of the source-level adaptation argued for above. As Table~\ref{tab:scope-comparison} shows, it is the only system that reaches the harness, covering a strict superset of the text-mutable scope addressed by prior work. MOSS surfaces its entire evolution lifecycle---triggering, status query, stop, apply, conversational flagging---to the agentic substrate as a built-in \texttt{moss evo} CLI capability via system-prompt injection, so the user interacts with evolution through the same conversational interface they already use for daily work. Evidence accumulates from two complementary channels: a periodic background scan over recent user sessions automatically surfaces under-performing dialogue segments, and the user can additionally flag any turn in conversation; both feed the same batch, and every downstream stage is anchored to fixing that batch rather than to a synthetic benchmark target. Code modification is carried out by a pluggable external coding-agent CLI invoked as a host-side subprocess, and candidates are verified on ephemeral trial workers that replay the batch against the candidate image in a production-equivalent container---not against unit tests or synthetic harnesses. Throughout the loop, every artifact---plans, diffs, build logs, per-iteration scoring matrices---is readable by the agent on demand, so the user can audit what the system intends to change and why before authorizing the swap. On convergence, MOSS proactively notifies the user through the same conversational channel and pauses; only after explicit user authorization does the host-daemon perform an in-place container swap, preserving user state and rolling back automatically on a failed post-swap health probe. On OpenClaw, MOSS lifts a four-task mean grader score from 0.25 to 0.61 in a single cycle, without human intervention.

\section{System Architecture}
\label{sec:arch}

MOSS realizes self-rewriting on a production agentic substrate through five pieces: a main container that runs the substrate's user-facing agent, a control-surface CLI through which the agent (and the user, via the agent) drives evolution, a pluggable external coding-agent CLI invoked per stage to make code edits, a host-resident daemon (the \emph{host-daemon}) that supervises everything else, and ephemeral trial workers used to verify candidates.

\subsection{Substrate}
\label{subsec:substrate}

For the exposition and case study throughout this paper, MOSS uses OpenClaw~\citep{openclaw-github} as its substrate; the design adapts to other mainstream agentic substrates such as Hermes Agent. OpenClaw is a production-grade agentic system, with a multi-channel gateway, plugin and hook plumbing, session and skill machinery, and persistent user state. The minimal scaffolds where academic self-evolving agents operate are far smaller and simpler; that gap shapes the four design choices below.

\subsection{Control Surface}
\label{subsec:control-surface}

MOSS exposes its entire evolution lifecycle to the agentic substrate through a single \texttt{moss evo} CLI that the substrate's user-facing agent invokes via its built-in shell tool. Nine subcommands cover the full surface: \texttt{status}, \texttt{batches}, \texttt{batch}, \texttt{start}, \texttt{stop}, \texttt{restart}, \texttt{apply}, \texttt{flag}, and \texttt{catch-up}. The first seven route over HTTP to an evolution-control endpoint group that MOSS injects into the substrate's gateway, where they reach the in-container evolution service; \texttt{flag} and \texttt{catch-up} route over a Unix socket to the host-daemon's auto-scan engine.

The CLI is reachable from inside the substrate's runtime through a single bind-mount of the script into the container at install time. To make the agent aware of the capability, MOSS injects a one-paragraph block into the substrate's system prompt pointing the agent at an on-disk capability document that describes the commands, their use, and the operational rules; the agent reads the document on demand when the user mentions evolution, batches, applying a new version, or expresses dissatisfaction with a turn. Asynchronous notifications from MOSS to the agent flow in the opposite direction through three webhook events: \texttt{evolution-converged} and \texttt{evolution-failed} are fired by the in-container evolution service at loop termination, and \texttt{apply-complete} is fired by the host-daemon after a container swap settles. Each event is POSTed to a webhook endpoint that a small hook-mapping configuration in the substrate translates into a system message delivered on the agent's next turn.

This three-piece interface---CLI in, webhook out, capability document as reference---is the entire contract between MOSS and a substrate. Any substrate that provides five primitives (shell-equivalent tool execution, filesystem read, periodic scheduling, webhook-to-agent delivery, and system-prompt injection) can host MOSS by configuring these pieces, with no changes to MOSS's own code.

\subsection{External Coding Agent}
\label{subsec:external-coding-agent}

On a substrate this large, MOSS cannot perform code modification through a single inference loop that reads and rewrites the codebase itself---the cross-file invariants and concurrent state interactions exceed what one such loop can carry. MOSS therefore handles evolution scheduling and decision-making through a deterministic state machine inside the main container, and delegates the specific act of editing to an external coding-agent CLI invoked as a host-side subprocess; the CLI runs on the host rather than inside the main container because only the host carries the shell environment, network egress, and scope mounts the editing work needs. Per-stage invocation mechanics are deferred to \S\ref{subsec:stage-pipeline}. Responsibilities split along this seam: MOSS owns stage ordering, verdicts, loop exit, and the swap moment, while the coding-agent CLI owns the act of editing within the scope it is given.

The coding-agent CLI itself is pluggable. MOSS abstracts the integration through a four-method runner interface; the dispatcher, RPC schema, stage logic, and substrate-side gateway are unchanged across providers. Four runners ship in tree---Claude Code~\citep{claude-code}, OpenAI Codex~\citep{openai-codex-cli}, DeepSeek-TUI~\citep{deepseek-tui}, and OpenCode~\citep{opencode}---selected at start time by configuration, with an optional per-spawn override that lets individual roles use different providers within one evolution run. Adding a fifth provider is one new runner file plus a single line in the registry. This decoupling keeps the evolution machinery independent of any specific coding-agent vendor and lets the substrate operator pair MOSS with whichever provider best matches their model-access and cost constraints.

\subsection{Host-Daemon Supervision}
\label{subsec:host-daemon}

Container lifecycle management cannot live inside the main container: once the candidate converges, the container must be stopped and replaced by a freshly built image, and a process cannot start a new container after exiting itself.

MOSS places that responsibility, and several related ones, in a host-resident asyncio process called the \emph{host-daemon}. It serves a Unix-socket RPC handling four families of operations: coding-agent CLI invocation per stage (\S\ref{subsec:external-coding-agent}), trial-worker container lifecycle, docker build and image management, and the auto-scan engine that scans session JSONLs for under-performing dialogue chunks (\S\ref{subsec:directed}). A swap-supervisor task runs in the same event loop, file-polling the evolution state directory for swap requests written by the gateway's apply handler; on detection, it restarts the substrate container against the candidate image, probes the new container's health, and either commits the swap or rolls back to the last-known-good image (the probe protocol and state-file layout are described in \S\ref{subsec:swap}). After the swap settles, the daemon fires the \texttt{apply-complete} webhook back to the freshly-swapped gateway so the new instance's agent receives a system message announcing what just happened.

\subsection{Topology}
\label{subsec:topology}

\begin{figure}[ht]
\centering
\begin{tikzpicture}[
  font=\small,
  >={Stealth[length=2mm]},
  box/.style={
    draw, thick, rounded corners=2pt, align=center,
    inner sep=5pt, minimum height=1.4cm, minimum width=4.0cm
  },
  hostbg/.style={
    draw, dashed, rounded corners=4pt, inner sep=10pt
  },
  edgelabel/.style={font=\scriptsize, fill=white, inner sep=1.2pt, align=center}
]

\node[box, fill=blue!7] (gateway) {%
  moss-gateway container\\[1pt]
  {\footnotesize agent + evolution service}\\[1pt]
  {\footnotesize + bind-mounted \texttt{moss} CLI}
};

\node[box, fill=orange!14, right=20mm of gateway] (daemon) {%
  host-daemon\\[1pt]
  {\footnotesize asyncio RPC server,}\\[1pt]
  {\footnotesize swap supervisor,}\\[1pt]
  {\footnotesize auto-scan engine}
};

\node[box, fill=green!10, below=14mm of daemon, xshift=-2cm] (cli) {%
  Coding-agent CLI\\[1pt]
  {\footnotesize spawned per stage}
};

\node[box, fill=purple!8, below=14mm of daemon, xshift=2cm] (trial) {%
  Trial workers $\times$\,$N$\\[1pt]
  {\footnotesize ephemeral, per iteration}
};

\draw[<->, thick] (gateway.north east) to[bend left=15] node[midway, edgelabel] {HTTP\\(api + hooks)} (daemon.north west);
\draw[<->, thick] (gateway.south east) to[bend right=15] node[midway, edgelabel] {Unix socket\\(RPC)} (daemon.south west);
\draw[->, thick]  (daemon) -- (cli) node[midway, left=1mm, edgelabel] {spawn};
\draw[->, thick]  (daemon) -- (trial) node[midway, right=1mm, edgelabel] {spawn};

\begin{scope}[on background layer]
\node[hostbg, fit=(daemon)(gateway)(cli)(trial), fill=gray!4] (host) {};
\end{scope}
\node[font=\footnotesize, anchor=north west] at ([xshift=4pt,yshift=-4pt]host.north west) {Host machine};

\end{tikzpicture}
\caption{MOSS host-side topology. The moss-gateway container hosts the user-facing agent, the in-container evolution service, and a bind-mounted \texttt{moss} CLI; the host-daemon (an asyncio process) serves a Unix-socket RPC for coding-agent and trial-worker orchestration, exposes an HTTP path for evolution control via the gateway, supervises swap requests, and runs the auto-scan engine. The coding-agent CLI is spawned per evolution stage; trial workers are launched per iteration as ephemeral containers. The user-state volume (sessions, memory, credentials, agent configs) is mounted into the moss-gateway container from the host filesystem so state survives an in-place swap (not shown).}
\label{fig:topology}
\end{figure}
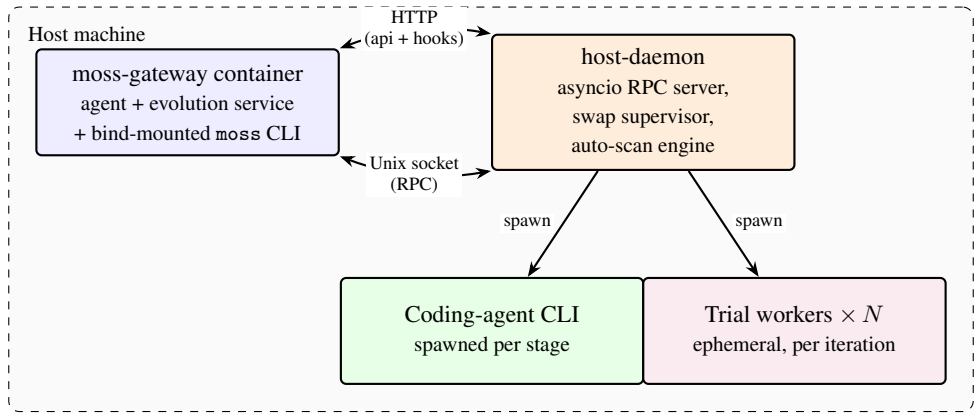

The four preceding choices materialize into the host-side topology in Figure~\ref{fig:topology}.

The host-daemon runs permanently on the host and supervises three other host-side entities. The moss-gateway container is the long-running, user-facing substrate instance hosting the chat agent, the in-container evolution service, and the bind-mounted \texttt{moss} CLI; its image tag is bumped on every successful swap. The coding-agent CLI is a subprocess the daemon spawns per evolution stage (\S\ref{sec:evolution}); it lives only for the duration of that stage. The trial workers are short-lived in a different sense: each iteration brings up $N$ containers from the candidate image (\S\ref{subsec:verification}), runs the agent against each batch task autonomously, then tears them down.

User state---sessions, memory, credentials, and agent configs---lives on a user-state volume mounted from the host filesystem into the moss-gateway container. A swap therefore destroys the container but not the volume, and the new image inherits state intact; trial workers run without this mount and with isolated networking.

\section{The Evolution Process}
\label{sec:evolution}

This section walks one full evolution of MOSS, from a curated failure batch to a swapped-in stronger agent: a directed goal from real user-session evidence (\S\ref{subsec:directed}), a bounded iteration loop on a baseline (\S\ref{subsec:loop}), the seven-stage pipeline per iteration (\S\ref{subsec:stage-pipeline}), runtime verification on the candidate image (\S\ref{subsec:verification}), and in-place container swap (\S\ref{subsec:swap}).

\subsection{Directed Evolution}
\label{subsec:directed}

In the academic line, source-level self-evolving systems follow an exploratory paradigm: each candidate is scored against a fixed benchmark, random mutations are then proposed on the basis of those scores, and the better-performing variants are retained for further iteration. This paradigm does not transfer to a production-grade agentic system. The codebase is large enough that random or undirected mutation rarely lands on any useful change; quality on a production agent is task-specific and admits no benchmark-style global fitness signal; under live user traffic any unanchored patch is indistinguishable from a regression; and users themselves expect the specific scenario they encountered to be fixed, not the agent globally retuned. MOSS therefore takes a directed and deterministic approach: each evolution is anchored to a concrete batch of production-failure evidence, and every downstream stage is bound to fixing precisely that batch.

Evidence enters the batch through two CLI-mediated paths that share the same backend. A periodic cron job runs \texttt{moss evo catch-up}, which scans every agent's session JSONLs from a per-session cursor and slices new content into chunks; inline, the Task-Evaluate stage scores each chunk's keypoints and only those tagged \texttt{weak} or \texttt{missing} are appended to that conversation's current open batch. In parallel, when the user expresses dissatisfaction in conversation, the agent invokes \texttt{moss evo flag}, which performs the same scan against a single session from its cursor to EOF and yields chunks the same way. Both paths append to the open batch of the source conversation---one open batch is maintained per conversation, sealed and replaced with a fresh open batch when its chunk count reaches a configurable threshold (default 8). Evolution is triggered by \texttt{moss evo start}; with no argument the latest non-empty batch is selected, and if it is still open it is sealed first.

\subsection{Evolution Loop}
\label{subsec:loop}

Single-shot patching at this scale is unreliable: localization can miss relevant files, an initial plan can misdiagnose root cause or mis-scope, and a first implementation can regress on adjacent behavior. MOSS therefore iterates. Each iteration's structured evaluation of the patched candidate feeds the next iteration's localization and planning, refining direction rather than retrying from scratch. Because directed evolution has a definable success condition---the batch is fixed---the loop is bounded: it exits when that condition is met or when further work is judged unproductive.

A baseline evaluation runs before iteration 1. The Task-Evaluate stage scores the original transcripts already attached to each batch chunk against a chosen keypoint set, producing a baseline keypoint matrix that anchors every subsequent comparison (\S\ref{subsec:verification}).

After baseline, each iteration runs the seven-stage pipeline (\S\ref{subsec:stage-pipeline}) and ends with one of four verdicts. Three are terminal: the batch is fixed (loop exits to \S\ref{subsec:swap}), the base model has hit its ceiling, or no patch within the current harness can resolve the failure. The fourth---progress was made but more work is needed---continues into the next iteration. A plateau guard backs this up: when no keypoint has improved for several consecutive iterations, the orchestrator forces convergence at the current peak rather than consuming further trials.

An evolution-depth dial---light, standard, or deep---scales the iteration budget. Maximum iteration count, per-stage round budgets, trials per task, and the plateau threshold all shift together. The default is standard; users facing more critical regressions can request deep.

\subsection{Stage Pipeline}
\label{subsec:stage-pipeline}

One iteration decomposes into seven stages. A single prompt asked to diagnose, plan, implement, verify, and decide overloads context and produces lower-quality output than a sequenced flow. Each stage also reads a different artifact---traces, diagnosis, diff, trial output, cross-iteration matrix---and gate-based quality control requires explicit boundaries: a review between planning and implementation only exists if planning is its own stage. The Task-Evaluate stage stands alone for a further reason: its output is archived and compared across iterations. MOSS executes these stages in sequence by invoking a coding-agent CLI (\S\ref{subsec:external-coding-agent}) once at each stage. Build and Trial (\S\ref{subsec:verification}) are runtime affordances around the Task-Evaluate stage rather than reasoning stages themselves, and so are not counted in the seven.

The seven stages, in order, are:

\begin{enumerate}
  \item \textbf{Locate} --- reads baseline traces and batch failures and writes a diagnosis without proposing fixes.
  \item \textbf{Plan} --- identifies root cause and specifies the fix: which files change, what logic is added, what is left alone.
  \item \textbf{Plan-Review} --- first quality gate; the plan is approved, rejected as architecturally off-target, or rejected as too narrow. The Plan and Plan-Review stages alternate within a multi-round plan-loop until approval or a round-budget cap.
  \item \textbf{Implement} --- writes the code in the inner substrate repository as a single git commit.
  \item \textbf{Code-Review} --- second quality gate; the diff is reviewed against the plan and approved or rejected. The Implement and Code-Review stages alternate within a multi-round code-loop, with the working tree hard-reset to the loop's starting commit between rounds.
  \item \textbf{Task-Evaluate} --- after Build and Trial (\S\ref{subsec:verification}), scores four to seven keypoints per task on a four-level qualitative scale (\texttt{strong} / \texttt{adequate} / \texttt{weak} / \texttt{missing}). The same stage runs once during pre-loop baseline (\S\ref{subsec:loop}); the keypoint set chosen there is locked for the rest of the loop.
  \item \textbf{Verdict} --- synthesizes all per-task evaluations and the cross-iteration keypoint matrix into one of four verdicts: \texttt{CONVERGED}, \texttt{NEED\_MORE\_WORK}, \texttt{FUNDAMENTAL\_LIMIT\_MODEL}, or \texttt{FUNDAMENTAL\_LIMIT\_ARCHITECTURE}. The orchestrator may downgrade \texttt{NEED\_MORE\_WORK} to \texttt{CONVERGED} if the keypoint matrix shows no improvement across a depth-dependent plateau window.
\end{enumerate}

\begin{figure}[ht]
\centering
\begin{tikzpicture}[
  font=\footnotesize,
  every node/.style={align=center},
  layer/.style={
    draw, thick, rounded corners=3pt, inner sep=6pt
  }
]

\node[layer, fill=blue!7, text width=12cm] (L0) {%
  Layer 0: pre-loop baseline
};

\node[layer, fill=orange!10, text width=12cm, below=4mm of L0] (L1) {%
  Layer 1: iteration loop\\[1pt]
  iter\,1 $\rightarrow$ iter\,2 $\rightarrow$ \ldots $\rightarrow$ iter\,$N$\\[6pt]
  \begin{tikzpicture}[font=\footnotesize, every node/.style={align=center}]
    \node[layer, fill=green!10, text width=10.6cm] (L2) {%
      Layer 2: stage pipeline\\[1pt]
      \textcircled{\footnotesize 1}\,Locate $\rightarrow$ \textcircled{\footnotesize 2}\,Plan $\rightarrow$ \textcircled{\footnotesize 3}\,Plan-Review $\rightarrow$
      \textcircled{\footnotesize 4}\,Implement $\rightarrow$\\
      \textcircled{\footnotesize 5}\,Code-Review $\rightarrow$ \textcircled{\footnotesize 6}\,Task-Evaluate $\rightarrow$ \textcircled{\footnotesize 7}\,Verdict\\[6pt]
      \begin{tikzpicture}[font=\footnotesize, every node/.style={align=center}]
        \node[layer, fill=purple!10, text width=9.2cm] {%
          Layer 3: stage-internal retry round\\[1pt]
          round\,0 $\rightarrow$ round\,1 $\rightarrow$ \ldots
        };
      \end{tikzpicture}
    };
  \end{tikzpicture}
};

\end{tikzpicture}
\caption{The four nested levels of MOSS evolution: a pre-loop baseline (Layer 0), an iteration loop (Layer 1), a fixed-order seven-stage pipeline per iteration (Layer 2), and internal multi-round loops around the two review gates (Layer 3).}
\label{fig:nesting}
\end{figure}
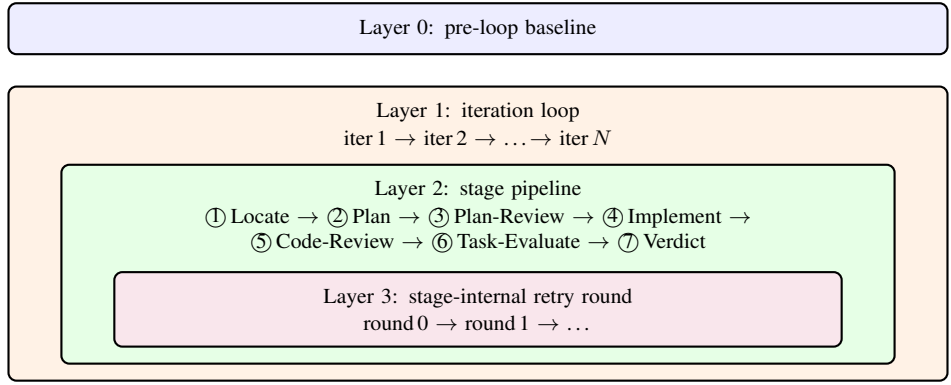

\subsection{Runtime Verification}
\label{subsec:verification}

The Code-Review stage operates at the syntactic and semantic level, so runtime faults pass through it: race conditions, cross-module state interactions, and hook-order-dependent behavior only manifest when the agent runs. Entire classes of application-level failure---channel routing, session lifecycle, cross-module state propagation---surface only under execution. Verification must therefore be runtime, on a production-equivalent environment, and against the same prompts that produced the failure evidence in the first place.

MOSS realizes this through ephemeral trial workers. After Build, the host-daemon spawns $N$ short-lived containers from the candidate image---the same image that would be promoted on convergence---and has the agent autonomously process the batch tasks inside each, repeating every task several trials to expose flakiness. Trial workers are network- and mount-isolated from the live main container and are torn down when the iteration ends. The pre-loop baseline does not require running the agent again: each batch chunk already contains the original transcript captured by auto-scan at evidence time, and the Task-Evaluate stage scores those transcripts directly to lock the keypoint set for the rest of the loop. This keeps verification closer to the deployed setting than gating on unit tests or external benchmarks would.

\subsection{In-Place Container Swap}
\label{subsec:swap}

Standard deployment patterns do not fit the production setting. The agent is a single-instance system bound to user state---sessions, memory, credentials, and agent configs all live in the user-state volume---so blue-green or canary would demand sticky-session routing or non-trivial state migration. User state must also \emph{survive} the swap so that the user experiences a better agent rather than a replaced one, a constraint minimal-scaffold self-evolving agents do not face. Rollback is mandatory because trial workers run against batch tasks in isolation, without the live container's user state or production traffic, and a candidate that passes trial can still regress live.

Convergence does not automatically promote the candidate. On the \texttt{CONVERGED} verdict, the evolution loop marks the batch as ready to apply and fires an \texttt{evolution-converged} webhook that the gateway routes to the agent as a system message; promotion requires the user to invoke \texttt{moss evo apply}, typically through the agent in conversation. The apply request HTTP-POSTs to the gateway, which atomically writes a swap-request file; the host-daemon's swap-supervisor file-polls that path every two seconds and, on detection, restarts the substrate container against the candidate image. A 90-second probe window follows, sampling every 5 seconds; each probe runs four health checks (heartbeat freshness $\leq 30$\,s, container running, and two substrate-level CLI status probes) and three consecutive passes commit the swap, otherwise the supervisor rolls back. The rollback target is read from an independent last-known-good image record rather than from the swap request itself, so a stale request cannot trap the system in a rollback loop. After either outcome the supervisor fires an \texttt{apply-complete} webhook---carrying success or rolled-back status---to the now-fresh gateway, where the agent receives it as a system message and re-orients to the new state. The user-state volume is mounted into the new container untouched, carrying sessions, memory, credentials, and agent configs across the swap.

\section{Case Study}
\label{sec:case-study}

This section presents a case study of MOSS's evolution loop, using a controlled batch of \texttt{claweval}~\citep{ye2026claw} benchmark tasks as input to demonstrate empirically that the loop produces an effective harness-level fix. We feed four compliance-audit tasks to the loop as the failure batch, run the pipeline end-to-end (pre-loop baseline $\rightarrow$ stage pipeline $\rightarrow$ trial verification $\rightarrow$ host-daemon swap), and measure the iteration-1 candidate image against the baseline image on the same four tasks. The score difference, together with the qualitative change in agent output, is the evidence that the loop closes the gap from a controlled failure signal to a deployed agent improvement.

\subsection{Setup: Tasks and Baseline}
\label{subsec:case-setup}

The batch consists of four \texttt{claweval} tasks in the operations / compliance-audit category. T141zh / T142 (SLA compliance audit, Chinese / English) asks the agent to list P1 violation tickets and compute per-ticket SLA compliance against tiered rules served by a config mock service. T137zh / T138 (restock-chain check, Chinese / English) asks the agent to trace a multi-service chain---scheduler jobs, integration configs, inventory levels---and report which restock jobs are failing, which integration is broken, and which inventory is short. Each pair shares the same task across two language variants. We use these four tasks both as the input batch fed into evolution and as the test set on which the post-evolution candidate is re-scored---a controlled setup that lets the same grader witness both ends of the loop.

The agent under evolution is OpenClaw with DeepSeek V3.2~\citep{liu2025deepseek} as its underlying model. Baseline runs score 0.21--0.33 with a mean near 0.25 on the \texttt{claweval} grader's $[0,1]$ scale, well below its 0.75 pass threshold. Baseline trial transcripts show recurring failure modes across both task families: on the SLA-compliance pair, the agent typically lists only some of the relevant tickets, marks the others as ``response data incomplete, compliance status indeterminate'', and occasionally misattributes customer names between adjacent tickets; on the restock-chain pair, the response is similarly partial, with broken or missing chains between scheduler jobs, integrations, and inventories. These are the failure modes the evolution loop is asked to fix.

The \texttt{claweval} grader serves only as an external numeric witness for the reader; it is separate from MOSS's internal Task-Evaluate stage, which emits the qualitative keypoint matrix that drives the verdict (\S\ref{subsec:stage-pipeline}). Using a benchmark batch here trades realism for reproducibility---running the same four tasks pre- and post-evolution provides a controlled measurement that an organic user session could not anchor.

\subsection{What MOSS Does With the Batch}
\label{subsec:case-diag}

Given this input batch, MOSS runs the pre-loop baseline: the Task-Evaluate stage scores pre-captured baseline transcripts for each of the four tasks into a baseline keypoint matrix that locks the keypoint set for the rest of the loop. The matrix comes back weak or missing on tool sequencing, information extraction, and result reporting.

The Locate stage ingests the baseline transcripts and identifies a coverage gap in the harness's tool-result handling for multi-tool execution patterns. The agent under evolution routinely chooses one of the harness's generic execution paths over the semantic tools the mediator was designed around, and the mediator has no annotation branch for that path. A secondary parsing issue in the harness's dispatch-synthesis pipeline compounds the gap when the agent batches several lookups into a single shell construct, leaving downstream consumers with merged, partially-attributed outputs. Together these two surfaces explain the half-answers visible in the baseline trial transcripts.

The Plan stage translates the diagnosis into a two-surface fix inside the harness: an additional annotation branch that injects an explicit usage hint when the execution path returns a multi-entity payload, and a pre-call deny gate that blocks the specific batched-shell pattern so the agent is steered toward separate, individually-parsable calls. The Plan-Review stage approves the design without revisions. The Implement stage lands it as a single commit on the inner OpenClaw repository, modifying three files with 177 insertions and 1 deletion: a new annotation branch and supporting helpers in the tool-result mediator, an added pre-call check in the before-tool-call hook chain, and a new mediator test file exercising both surfaces. The change touches the harness itself rather than any text-mutable artifact, locating the fix where \S\ref{sec:intro} argues prior systems cannot reach.

\begin{figure}[ht]
\centering
\begin{tikzpicture}[
  font=\footnotesize,
  >={Stealth[length=2mm]},
  pipestep/.style={
    draw, thick, rounded corners=2pt, fill=blue!6,
    minimum width=2.0cm, minimum height=0.8cm, align=center
  }
]

\node[pipestep] (s1) {\textcircled{\scriptsize 1}\,Locate};
\node[pipestep, right=4mm of s1] (s2) {\textcircled{\scriptsize 2}\,Plan};
\node[pipestep, right=4mm of s2] (s4) {\textcircled{\scriptsize 4}\,Implement};
\node[pipestep, fill=orange!12, right=4mm of s4] (build) {Build};

\node[pipestep, fill=purple!10, below=9mm of s1] (trial) {Trial};
\node[pipestep, right=4mm of trial] (s6) {\textcircled{\scriptsize 6}\,Task-Evaluate};
\node[pipestep, right=4mm of s6] (s7) {\textcircled{\scriptsize 7}\,Verdict};

\draw[->, thick] (s1) -- (s2);
\draw[->, thick] (s2) -- (s4);
\draw[->, thick] (s4) -- (build);

\draw[->, thick] (build.south) -- ++(0,-3mm) -| (trial.north);

\draw[->, thick] (trial) -- (s6);
\draw[->, thick] (s6) -- (s7);

\end{tikzpicture}
\caption{Iteration-1 trace covering all stages MOSS executed; stage 3 (Plan-Review) and stage 5 (Code-Review) gates are elided for compactness.}
\label{fig:iter1-timeline}
\end{figure}
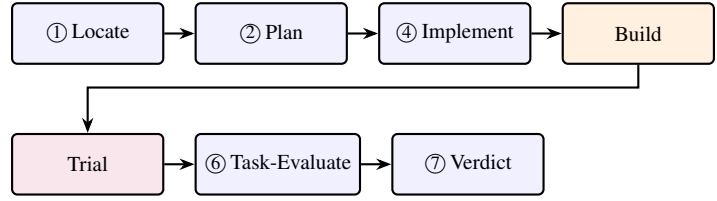

The Code-Review stage approves, the host-daemon builds the candidate image, and trial workers replay the batch against it. Figure~\ref{fig:iter1-timeline} sketches the resulting iteration-1 timeline. The Task-Evaluate stage rescores the transcripts; the Verdict stage compares the new keypoint matrix against the baseline, observes broad lift across tool-sequencing and result-reporting keypoints, and the candidate converges.

\subsection{Results: Iteration-1 Outcome}
\label{subsec:case-results}

On convergence, the host-daemon performs the in-place container swap described in \S\ref{subsec:swap}; for this benchmark run, the user-consent gate is auto-acknowledged so the evaluation pipeline can complete without interactive review. The converged candidate image is then re-run against the same four tasks, with grader scores reported in Table~\ref{tab:case-results}.

\begin{table}[ht]
\caption{Per-task \texttt{claweval} grader scores (mean of 3 trials per task) before and after the iteration~1 swap.}
\label{tab:case-results}
\centering
\begin{tabular}{lccc}
\toprule
Task & Baseline & Iter 1 & $\Delta$ \\
\midrule
T141zh\_sla\_compliance\_audit  & 0.3273          & 0.5330          & $+0.2057$ \\
T142\_sla\_compliance\_audit    & 0.2527          & 0.5453          & $+0.2926$ \\
T137zh\_restock\_chain\_check   & 0.2213          & 0.4567          & $+0.2354$ \\
T138\_restock\_chain\_check     & 0.2090          & 0.9049          & $+0.6959$ \\
\midrule
\textbf{mean}                    & \textbf{0.2526} & \textbf{0.6100} & $\boldsymbol{+0.3574}$ \\
\bottomrule
\end{tabular}
\end{table}

The four-task mean rises from 0.2526 to 0.6100. The largest gain is on T138, which climbs to 0.9049 with all three trials clearing the 0.75 pass threshold. The SLA pair (T141zh / T142) moves from the 0.25--0.33 range to roughly 0.53--0.55: the annotation path closes the largest defect, while per-ticket time-difference arithmetic and SLA-tier classification carry their own difficulty that one iteration does not resolve. T137zh shows the smallest gain at $+0.2354$, with one trial reaching pass and two below threshold.

The scores correspond to the agent's actual outputs. The SLA-compliance task transcripts that previously came back as the half-answers described in \S\ref{subsec:case-setup} now return per-ticket SLA-tier classifications and an aggregate summary (``2 of 6 P1 tickets violated SLA, affecting Customer B and Customer D, mean overrun 2.3h''); the restock-chain transcripts likewise return complete chains. The model and the task definition are unchanged; the harness change is the proximate cause.

\section{Related Work}
\label{sec:related}

\subsection{Agentic Systems}
\label{subsec:related-agentic}

Capable conversational models~\citep{achiam2023gpt} reframed the LLM as a system component rather than a demo, which opened the question of how to let it act. Tool-augmented language models answered first: Toolformer~\citep{schick2023toolformer} showed an LLM can learn when to call an API, and ToolLLM~\citep{qin2024toolllm} scaled the idea to thousands of real-world tools, turning the model from a text generator into a callable executor. ReAct~\citep{yao2022react} closed that executor into a loop by interleaving chain-of-thought reasoning with environment actions, and the resulting think/act cycle has since become the inner core of every agent that followed. Once one agent worked, splitting roles across many agents within a single inference frame became natural---MetaGPT~\citep{hong2024metagpt}, AutoGen~\citep{wu2024autogen}, ChatDev~\citep{qian2024chatdev}, and CAMEL~\citep{li2023camel} each demonstrated different forms of role specialization---and the design space has by now been mapped at survey scale~\citep{wang2024survey,guo2024large}. This lineage now extends to application-level production agents: Anthropic's Claude assistants~\citep{claude3-anthropic2024} and the open-source OpenClaw~\citep{openclaw-github} run as complete production-grade agentic systems---multi-channel, multi-user services with persistent state and a full harness layer governing routing, gating, and lifecycle. 
\subsection{Self-Evolving Agents}
\label{subsec:related-evolving}

Research on self-evolving agents has developed along two lines. The academic line establishes source-level self-evolution as a working primitive on minimal scaffolds. SICA~\citep{robeyns2025self} shows an agent can edit its own implementation and lift its own SWE-Bench score, settling the feasibility question. The Darwin G\"{o}del Machine~\citep{zhang2025darwin} reframes the same primitive as open-ended search over an archive of agent variants; HyperAgents~\citep{zhang2026hyperagents} pushes that style further by making the meta-procedure itself editable. Meta-Harness~\citep{lee2026meta} isolates a finer signal: exposing past execution traces to the coding-agent proposer strengthens iterative improvement beyond what benchmark-score feedback alone yields. The application-level line extends evolution to deployed agents but bounds its scope to text-mutable artifacts. Hermes Agent~\citep{hermes-github}, paired with the DSPy~\citep{khattab2023dspy} and GEPA~\citep{agrawal2025gepa} optimizers that compile and search over prompts, forms the most ambitious cluster on this side; Capability Evolver~\citep{wang2026procedural}, SkillClaw~\citep{ma2026skillclaw}, GenericAgent~\citep{liang2026genericagent}, and EvoAgentX~\citep{wang2025evoagentx} each occupy a different text-mutable layer---behavioral genes and memory capsules, shared skill corpora, markdown SOPs, prompts and workflow graphs---and none touches the harness. MOSS combines the two: source-level evolution applied to an application-level substrate, with the harness layer included and deployment-failure signals replacing benchmark scores.

\section{Conclusion}
\label{sec:conclusion}

We have argued that source-level adaptation is a fundamentally more general medium for self-evolving agents than the text-mutable scope to which prior application-level systems are confined: it is Turing-complete, a strict superset of every text-mutable design space, deterministic in effect, and stable under long-context drift. MOSS instantiates this argument on a production agentic substrate, extending the editable scope from skill files, prompt configurations, memory schemas, and workflow graphs to the agent harness itself, where routing, state management, hook ordering, and dispatch live. Anchoring evolution to an automatically curated batch of production-failure evidence, executing a deterministic multi-stage pipeline with code modification delegated to a pluggable external coding-agent CLI, verifying candidates by replaying the batch against the candidate image in ephemeral trial workers, and promoting them via user-consent-gated, in-place container swap with health-probe-gated rollback together close the loop from a real user failure to a deployed harness-level fix; on OpenClaw, this single loop lifts a four-task mean grader score from 0.25 to 0.61 without human intervention.

\bibliographystyle{plainnat}
\bibliography{references}

@inproceedings{hu2025automated,
  title={Automated design of agentic systems},
  author={Hu, Shengran and Lu, Cong and Clune, Jeff},
  booktitle={International Conference on Learning Representations},
  volume={2025},
  pages={21344--21377},
  year={2025}
}

@article{robeyns2025self,
  title={A self-improving coding agent},
  author={Robeyns, Maxime and Szummer, Martin and Aitchison, Laurence},
  journal={arXiv preprint arXiv:2504.15228},
  year={2025}
}

@misc{openclaw-github,
  title        = {{OpenClaw --- Personal AI Assistant}},
  author       = {{OpenClaw}},
  year         = {2024},
  howpublished = {GitHub repository},
  note         = {Open-source multi-channel agentic system; supports WhatsApp, Telegram, Slack, Discord, Google Chat, Signal, Microsoft Teams, Matrix, Feishu, and other channels.},
  url          = {https://github.com/openclaw/openclaw}
}

@misc{hermes-github,
  title        = {{Hermes Agent: The self-improving AI agent built by Nous Research}},
  author       = {{Nous Research}},
  year         = {2024},
  howpublished = {GitHub repository},
  note         = {Application-level self-evolving agent with a built-in learning loop: autonomous skill creation after complex tasks, skills self-improve during use, agent-curated memory, and procedural memory.},
  url          = {https://github.com/NousResearch/hermes-agent}
}

@misc{hermes-self-evo,
  title        = {{Hermes Agent Self-Evolution: Evolutionary self-improvement for Hermes Agent}},
  author       = {{Nous Research}},
  year         = {2026},
  howpublished = {GitHub repository},
  note         = {Companion project to Hermes Agent that evolves skills, tool descriptions, system prompts, and code via DSPy + GEPA; gates every variant on full pytest pass and size limits (skills $\leq$ 15KB, tool descriptions $\leq$ 500 chars).},
  url          = {https://github.com/NousResearch/hermes-agent-self-evolution}
}

@misc{claude3-anthropic2024,
  title        = {{The Claude 3 Model Family: Opus, Sonnet, Haiku}},
  author       = {{Anthropic}},
  year         = {2024},
  month        = mar,
  howpublished = {Anthropic technical report},
  note         = {Model card for the Claude 3 family released March 2024; documents capabilities, benchmarks, and safety evaluations for Opus, Sonnet, and Haiku.},
  url          = {https://www.anthropic.com/claude-3-model-card}
}

@article{wang2026procedural,
  title={From Procedural Skills to Strategy Genes: Towards Experience-Driven Test-Time Evolution},
  author={Wang, Junjie and Ren, Yiming and Zhang, Haoyang},
  journal={arXiv preprint arXiv:2604.15097},
  year={2026}
}

@article{ma2026skillclaw,
  title={Skillclaw: Let skills evolve collectively with agentic evolver},
  author={Ma, Ziyu and Yang, Shidong and Ji, Yuxiang and Wang, Xucong and Wang, Yong and Hu, Yiming and Huang, Tongwen and Chu, Xiangxiang},
  journal={arXiv preprint arXiv:2604.08377},
  year={2026}
}

@article{liang2026genericagent,
  title={GenericAgent: A Token-Efficient Self-Evolving LLM Agent via Contextual Information Density Maximization (V1. 0)},
  author={Liang, Jiaqing and Han, Jinyi and Li, Weijia and Wang, Xinyi and Zhang, Zhoujia and Jiang, Zishang and Liao, Ying and Li, Tingyun and Huang, Ying and Shen, Hao and others},
  journal={arXiv preprint arXiv:2604.17091},
  year={2026}
}

@inproceedings{wang2025evoagentx,
  title={Evoagentx: An automated framework for evolving agentic workflows},
  author={Wang, Yingxu and Liu, Siwei and Fang, Jinyuan and Meng, Zaiqiao},
  booktitle={Proceedings of the 2025 Conference on Empirical Methods in Natural Language Processing: System Demonstrations},
  pages={643--655},
  year={2025}
}

@article{zhang2025darwin,
  title={Darwin godel machine: Open-ended evolution of self-improving agents},
  author={Zhang, Jenny and Hu, Shengran and Lu, Cong and Lange, Robert and Clune, Jeff},
  journal={arXiv preprint arXiv:2505.22954},
  year={2025}
}

@article{zhang2026hyperagents,
  title={Hyperagents},
  author={Zhang, Jenny and Zhao, Bingchen and Yang, Wannan and Foerster, Jakob and Clune, Jeff and Jiang, Minqi and Devlin, Sam and Shavrina, Tatiana},
  journal={arXiv preprint arXiv:2603.19461},
  year={2026}
}

@article{lee2026meta,
  title={Meta-harness: End-to-end optimization of model harnesses},
  author={Lee, Yoonho and Nair, Roshen and Zhang, Qizheng and Lee, Kangwook and Khattab, Omar and Finn, Chelsea},
  journal={arXiv preprint arXiv:2603.28052},
  year={2026}
}

@article{ye2026claw,
  title={Claw-eval: Toward trustworthy evaluation of autonomous agents},
  author={Ye, Bowen and Li, Rang and Yang, Qibin and Liu, Yuanxin and Yao, Linli and Lv, Hanglong and Xie, Zhihui and An, Chenxin and Li, Lei and Kong, Lingpeng and others},
  journal={arXiv preprint arXiv:2604.06132},
  year={2026}
}

@article{liu2025deepseek,
  title={Deepseek-v3. 2: Pushing the frontier of open large language models},
  author={Liu, Aixin and Mei, Aoxue and Lin, Bangcai and Xue, Bing and Wang, Bingxuan and Xu, Bingzheng and Wu, Bochao and Zhang, Bowei and Lin, Chaofan and Dong, Chen and others},
  journal={arXiv preprint arXiv:2512.02556},
  year={2025}
}

@article{achiam2023gpt,
  title={Gpt-4 technical report},
  author={Achiam, Josh and Adler, Steven and Agarwal, Sandhini and Ahmad, Lama and Akkaya, Ilge and Aleman, Florencia Leoni and Almeida, Diogo and Altenschmidt, Janko and Altman, Sam and Anadkat, Shyamal and others},
  journal={arXiv preprint arXiv:2303.08774},
  year={2023}
}

@article{schick2023toolformer,
  title={Toolformer: Language models can teach themselves to use tools},
  author={Schick, Timo and Dwivedi-Yu, Jane and Dess{\`\i}, Roberto and Raileanu, Roberta and Lomeli, Maria and Hambro, Eric and Zettlemoyer, Luke and Cancedda, Nicola and Scialom, Thomas},
  journal={Advances in neural information processing systems},
  volume={36},
  pages={68539--68551},
  year={2023}
}

@inproceedings{qin2024toolllm,
  title={Toolllm: Facilitating large language models to master 16000+ real-world apis},
  author={Qin, Yujia and Liang, Shihao and Ye, Yining and Zhu, Kunlun and Yan, Lan and Lu, Yaxi and Lin, Yankai and Cong, Xin and Tang, Xiangru and Qian, Bill and others},
  booktitle={International Conference on Learning Representations},
  volume={2024},
  pages={9695--9717},
  year={2024}
}

@article{yao2022react,
  title={React: Synergizing reasoning and acting in language models},
  author={Yao, Shunyu and Zhao, Jeffrey and Yu, Dian and Du, Nan and Shafran, Izhak and Narasimhan, Karthik and Cao, Yuan},
  journal={arXiv preprint arXiv:2210.03629},
  year={2022}
}

@inproceedings{hong2024metagpt,
  title={MetaGPT: Meta programming for a multi-agent collaborative framework},
  author={Hong, Sirui and Zhuge, Mingchen and Chen, Jonathan and Zheng, Xiawu and Cheng, Yuheng and Wang, Jinlin and Zhang, Ceyao and Yau, Steven and Lin, Zijuan and Zhou, Liyang and others},
  booktitle={International Conference on Learning Representations},
  volume={2024},
  pages={23247--23275},
  year={2024}
}

@inproceedings{wu2024autogen,
  title={Autogen: Enabling next-gen LLM applications via multi-agent conversations},
  author={Wu, Qingyun and Bansal, Gagan and Zhang, Jieyu and Wu, Yiran and Li, Beibin and Zhu, Erkang and Jiang, Li and Zhang, Xiaoyun and Zhang, Shaokun and Liu, Jiale and others},
  booktitle={First conference on language modeling},
  year={2024}
}

@inproceedings{qian2024chatdev,
  title={Chatdev: Communicative agents for software development},
  author={Qian, Chen and Liu, Wei and Liu, Hongzhang and Chen, Nuo and Dang, Yufan and Li, Jiahao and Yang, Cheng and Chen, Weize and Su, Yusheng and Cong, Xin and others},
  booktitle={Proceedings of the 62nd annual meeting of the association for computational linguistics (volume 1: Long papers)},
  pages={15174--15186},
  year={2024}
}

@article{li2023camel,
  title={Camel: Communicative agents for" mind" exploration of large language model society},
  author={Li, Guohao and Hammoud, Hasan and Itani, Hani and Khizbullin, Dmitrii and Ghanem, Bernard},
  journal={Advances in neural information processing systems},
  volume={36},
  pages={51991--52008},
  year={2023}
}

@article{wang2024survey,
  title={A survey on large language model based autonomous agents},
  author={Wang, Lei and Ma, Chen and Feng, Xueyang and Zhang, Zeyu and Yang, Hao and Zhang, Jingsen and Chen, Zhiyuan and Tang, Jiakai and Chen, Xu and Lin, Yankai and others},
  journal={Frontiers of Computer Science},
  volume={18},
  number={6},
  pages={186345},
  year={2024},
  publisher={Springer}
}

@article{guo2024large,
  title={Large language model based multi-agents: A survey of progress and challenges},
  author={Guo, Taicheng and Chen, Xiuying and Wang, Yaqi and Chang, Ruidi and Pei, Shichao and Chawla, Nitesh V and Wiest, Olaf and Zhang, Xiangliang},
  journal={arXiv preprint arXiv:2402.01680},
  year={2024}
}

@article{khattab2023dspy,
  title={Dspy: Compiling declarative language model calls into self-improving pipelines},
  author={Khattab, Omar and Singhvi, Arnav and Maheshwari, Paridhi and Zhang, Zhiyuan and Santhanam, Keshav and Vardhamanan, Sri and Haq, Saiful and Sharma, Ashutosh and Joshi, Thomas T and Moazam, Hanna and others},
  journal={arXiv preprint arXiv:2310.03714},
  year={2023}
}

@article{agrawal2025gepa,
  title={Gepa: Reflective prompt evolution can outperform reinforcement learning},
  author={Agrawal, Lakshya A and Tan, Shangyin and Soylu, Dilara and Ziems, Noah and Khare, Rishi and Opsahl-Ong, Krista and Singhvi, Arnav and Shandilya, Herumb and Ryan, Michael J and Jiang, Meng and others},
  journal={arXiv preprint arXiv:2507.19457},
  year={2025}
}

@misc{claude-code,
  title        = {{Claude Code}},
  author       = {{Anthropic}},
  year         = {2025},
  howpublished = {\url{https://www.anthropic.com/claude-code}},
  note         = {Anthropic's command-line coding agent for the Claude model family.}
}

@misc{openai-codex-cli,
  title        = {{OpenAI Codex CLI}},
  author       = {{OpenAI}},
  year         = {2025},
  howpublished = {\url{https://github.com/openai/codex}},
  note         = {Open-source command-line coding agent backed by OpenAI models.}
}

@misc{deepseek-tui,
  title        = {{DeepSeek-TUI: Terminal UI for DeepSeek}},
  author       = {Hmbown},
  year         = {2025},
  howpublished = {\url{https://github.com/Hmbown/DeepSeek-TUI}},
  note         = {Terminal-UI coding agent for DeepSeek models.}
}

@misc{opencode,
  title        = {{OpenCode}},
  author       = {{OpenCode project}},
  year         = {2025},
  howpublished = {\url{https://opencode.ai}},
  note         = {Multi-provider open-source coding-agent CLI.}
}

\end{document}